\documentclass[sigconf]{acmart}

\usepackage{natbib}
 \usepackage{graphicx,graphics}
 \usepackage{amsmath}
 \usepackage{float}
\usepackage{algorithmicx,algorithm}
\usepackage{float}

\usepackage{nccmath}  

\newcounter{parentalgorithm}

\def\P{\mathbb{P}}
\def\E{\mathbb{E}}
\def\R{\mathbb{R}}

\def\P{\mathbb{P}}

\usepackage{subcaption}
\usepackage{enumitem}

\usepackage{xcolor}

\newtheorem{assumption}{Assumption}

\usepackage{algpseudocode}

\AtBeginDocument{%
  }

\copyrightyear{2026}
\acmYear{2026}
\setcopyright{cc}
\setcctype{by}
\acmConference[WWW '26]{Proceedings of the ACM Web Conference 2026}{April 13--17, 2026}{Dubai, United Arab Emirates}
\acmBooktitle{Proceedings of the ACM Web Conference 2026 (WWW '26), April 13--17, 2026, Dubai, United Arab Emirates}
\acmPrice{}
\acmDOI{10.1145/3774904.3792109}
\acmISBN{979-8-4007-2307-0/2026/04}

\begin{document}

\title{Auto-Bidding under Return-on-Spend Constraints with Uncertainty Quantification}


\settopmatter{authorsperrow=4}
\author{Jiale Han\textsuperscript{\textasteriskcentered}}
\affiliation{%
  \institution{University of California, Los Angeles}
  \city{Los Angeles}
  \country{USA}}
  \thanks{\textsuperscript{\textasteriskcentered} Work done during an internship at JD.com.}
\email{jialehan@ucla.edu}

\author{Chun Gan}
\affiliation{%
  \institution{JD.com}
  \city{Beijing}
  \country{China}
}
\email{ganchun1@jd.com}

\author{Chengcheng Zhang\textsuperscript{\textasteriskcentered}}
\affiliation{%
 \institution{The Chinese University of Hong Kong, Shenzhen}
 \city{Shenzhen}
 \country{China}}
\email{224010028@link.cuhk.edu.cn}

\author{Jie He}
\affiliation{%
  \institution{JD.com}
  \city{Beijing}
  \country{China}
}
\email{hejie67@jd.com}

\author{Zhangang Lin}
\affiliation{%
  \institution{JD.com}
  \city{Beijing}
  \country{China}
}
\email{linzhangang@jd.com}

\author{Ching Law}
\affiliation{%
  \institution{JD.com}
  \city{Beijing}
  \country{China}
}
\email{lawching@jd.com}

\author{Xiaowu Dai\textsuperscript{$\dagger$}}
\affiliation{%
  \institution{University of California, Los Angeles}
  \city{Los Angeles}
  \country{USA}
  \thanks{\textsuperscript{$\dagger$}Corresponding author.}}
\email{daix@ucla.edu}

\renewcommand{\shortauthors}{Jiale Han et al.}

\begin{abstract}
Auto-bidding systems are widely used in advertising to automatically determine bid values under constraints such as total budget and Return-on-Spend (RoS) targets. Existing works often assume that the value of an ad impression, such as the conversion rate, is known. This paper considers the more realistic scenario where the true value is unknown. We propose a novel method that uses conformal prediction to quantify the uncertainty of these values based on machine learning methods trained on historical bidding data with contextual features, without assuming the data are i.i.d. This approach is compatible with current industry systems that use machine learning to predict values. Building on prediction intervals, we introduce an adjusted value estimator derived from machine learning predictions, and show that it provides performance guarantees without requiring knowledge of the true value. We apply this method to enhance existing auto-bidding algorithms with budget and RoS constraints, and establish theoretical guarantees for achieving high reward while keeping RoS violations low. Empirical results on both simulated and real-world industrial datasets demonstrate that our approach improves performance while maintaining computational efficiency.
\end{abstract}

\ccsdesc[500]{Applied computing~Online auctions}
\keywords{Auto-bidding, Online Advertising, Uncertainty Quantification}


\maketitle
\newcommand\webconfavailabilityurl{https://doi.org/10.5281/zenodo.18308975}
\ifdefempty{\webconfavailabilityurl}{}{
\begingroup\small\noindent\raggedright\textbf{Resource Availability:}\\
The source code of this paper has been made publicly available at \url{\webconfavailabilityurl}.
\endgroup
}

\section{Introduction}
With the rapid digitization of commerce, online advertising has grown into a multi-billion-dollar industry. On advertising platforms, advertisers participate in real-time auctions triggered by user queries, competing for ad slots by submitting bids. The winner secures the slot and pays the platform. At the scale of trillions of auctions, traditional bidding methods where advertisers manually adjust bids for each query are infeasible. To address this challenge, ad platforms have developed auto-bidding services that place bids on behalf of advertisers, aiming to maximize returns while satisfying budget and constraints such as the return-on-spend (RoS) target, which requires the ratio of total value obtained by the advertiser to the total payment to exceed a specified minimum threshold, or other equivalent constraints, such as the target cost-per-acquisition (tCPA), which is the inverse of the target RoS \citep{zhang2014optimal, balseiro2019learning, aggarwal2019autobidding, he2021unified, balseiro2021robust, feng2023online, guo2024generative}.

Most existing auto-bidding algorithms assume that a bidder’s value for each ad impression is known. In practice, however, values can only be estimated using machine learning methods \citep{deng2020calibrating, yang2024deep}, and these predictions are inevitably uncertain due to market noise and model fitting \citep{hartline2013mechanism}.
Such uncertainty can significantly degrade performance compared with the idealized setting of perfect value knowledge, yet it is often overlooked. To address this gap, \citet{vijayan2025online} propose a UCB-based algorithm that learns value uncertainty and provides reward guarantees under budget and RoS constraints. However, like many other approaches offering reward guarantees \citep[e.g.,][]{feng2023online}, their method assumes identically and independently distributed (i.i.d.) data, even though bidders often face heterogeneous value distributions. In addition, it does not leverage machine learning predictions, assumes values are revealed immediately after winning, and requires solving a linear program for each auction, which is impractical in real-world systems where feedback is delayed \citep{deng2020calibrating} and computational cost is high.
These limitations motivate our central research question:

\indent
    \textit{Can machine learning predictions, despite noise and heterogeneity, be used to design auto-bidding algorithms that guarantee performance without knowing true values?}

To address this question, we propose a novel approach to characterize the uncertainty of values by employing the conformal prediction method \citep{barber2023conformal}.
Unlike the traditional conformal setting, we employ a new conformity score to provide a more precise upper bound for the values.
Our method builds upon machine learning predictions of the values using the contextual features, making it compatible with current auto-bidding systems. 
It does not require the data to be i.i.d.  and provides prediction intervals with a coverage guarantee under mild assumptions that hold in the online advertising setting.
We focus on the bidding problem of a single advertiser with budget and RoS constraints. We show that using the upper bounds of the prediction intervals of the values as a surrogate for the true values yields reward and RoS performance comparable to that achieved when using the true values.
In our method, the upper bound of the prediction interval is represented as the machine learning prediction plus an adjustment term, which can be interpreted as the adjusted predicted value that mitigates prediction bias.

We apply our method to design an adjusted online bidding algorithm that incorporates the adjusted predicted value into the online mirror descent framework of \citet{feng2023online}. Unlike their approach, which assumes perfect value knowledge and i.i.d. data, our method relaxes these assumptions while still achieving near-optimal bounds: a competitive ratio against the optimal reward and tight bounds on RoS violations.
To evaluate the efficiency of our method, we conduct extensive experiments comparing the adjusted algorithm with both the current algorithm and the state-of-the-art method from \citet{vijayan2025online}, which achieves near-optimal bounds on regret and constraint violation in settings where item values are unknown.  We perform simulations using a public dataset from Alibaba \citep{su2024auctionnet} and a real-world application with an industrial dataset from an e-commerce platform. Our method consistently outperforms all other algorithms and, compared to \citet{vijayan2025online}, significantly reduces computational time while achieving better performance in terms of reward and RoS violation.
In conclusion, our contributions are threefold.

(i) We propose an efficient machine learning–based approach to quantify value uncertainty from contextual features and historical data, without relying on traditional i.i.d. assumptions.  Building on this, we introduce an adjusted predicted value defined as the upper bound of the quantified uncertainty interval.

(ii) We establish theoretical guarantees on both reward and RoS when the adjusted predicted values are used as surrogates for the true values. Building on the algorithm proposed by \citet{feng2023online}, we derive new theoretical guarantees in the context of non-identically distributed requests, without requiring knowing the value.

(iii) Empirical evaluations on both simulated datasets from Alibaba and real-world industrial datasets demonstrate the effectiveness of our method, achieving higher performance in terms of reward and RoS violation while maintaining low computational cost.

\subsection{Related Work} 

\noindent
\textbf{Auto-bidding with constraints.} 
A growing body of work investigated bidding strategies in repeated auctions under value maximization with constraints \citep{aggarwal2019autobidding, he2021unified, feng2023online, guo2024generative, aggarwal2025no}. \citet{aggarwal2019autobidding} provided an optimal bidding framework for truthful auctions, which was later extended by \citet{he2021unified} to a framework using reinforcement learning. Building on this line of work, \citet{guo2024generative} employed this framework to develop a diffusion-based bidding algorithm. However, these approaches lacked strong theoretical guarantees on reward performance. Among the works that provided such guarantees, \citet{aggarwal2025no} proposed no-regret algorithms for non-truthful auctions against the best Lipschitz-continuous bidding strategy. \citet{feng2023online} established a no-regret algorithm for truthful auctions against the offline optimum. Nevertheless, these works assumed that item values were known prior to the auction. In contrast, our paper removed this assumption by incorporating uncertainty quantification, adapting the algorithm in \citet{feng2023online} to settings with unknown values and non-i.i.d. data.

\noindent
\textbf{Autobidding without knowing the values.}
Recent research has explored autobidding mechanisms without knowing the values \citep{weed2016online, achddou2021efficient, feng2018learning, castiglioni2022unifying, bernasconi2024beyond, vijayan2025online}. However, \citet{weed2016online}, \citet{achddou2021efficient}, and \citet{feng2018learning} primarily focused on the unconstrained setting, aiming to maximize utility, which was defined as the difference between the received value and the paid price.
In contrast, \citet{castiglioni2022unifying}, \citet{bernasconi2024beyond}, and \citet{vijayan2025online} investigated value maximization in constrained settings, assuming that bids could only take values from a finite set $\mathcal{B}$. However, the regret and constraint violation bounds derived in \citet{castiglioni2022unifying} and \citet{bernasconi2024beyond} exhibited a dependence of $O(\sqrt{|\mathcal{B}|})$.
While \citet{vijayan2025online} achieved near-optimal regret against the baseline by treating expectations as the benchmark, they assumed that auction requests were i.i.d. and that values were directly observed after winning the auction. These assumptions, however, were impractical in real-world online advertising, where the distribution of values was diverse, and feedback was often delayed \citep{deng2020calibrating}.
In contrast, our work did not need the assumptions of a finite set of bids and relaxed i.i.d. distributions, providing a more practical framework.

\noindent
\textbf{Value prediction in online advertising.} Value prediction was a fundamental problem in online advertising, where common target metrics included click-through rate (CTR) and conversion rate (CVR). Recent work has sought to improve value estimation by leveraging carefully engineered features and post-hoc user feedback \citep[e.g.,][]{deng2020calibrating, pan2020field, wei2022posterior, yang2024deep}. In particular, \citet{deng2020calibrating} proposed incorporating post-click signals—such as landing page session duration and the number of post-click page views—as features for value prediction, and then calibrated the predicted values using smoothed isotonic regression. \citet{pan2020field, wei2022posterior, yang2024deep} developed field-level calibration strategies, where features were classified into different fields and calibration was performed via neural networks. While these approaches provided effective means to improve calibration, a gap still remained between the predicted value and the realized post-hoc value due to inherent market uncertainty. Our work addressed this gap by offering rigorous uncertainty quantification for each post-hoc value prior to the auction, applicable to the current prediction system. This allowed for the construction of adjusted predicted values that provided formal theoretical performance guarantees for the bidding agent.

\section{Problem Setting}

\subsection{Auto-Bidding Problem}
We consider an auction setting with multiple bidders, each employing an automated bidding agent to participate in a sequence of auctions over a fixed time horizon—e.g., one day—with a total budget and desired RoS targets. 
We study the online auto-bidding problem from the perspective of a single bidder (bidding agent), who joins $T$ auctions in a time period.
The advertiser aims to maximize his own total values, subject to the budget constraint. 
In the $t$-th auction, $t\in[T]:=\{1, \dots, T\}$, let $b_t \in \R_{\geq0}$ denote the bidder’s bid, $v_t\in[0,1]$ represent the bidder’s value for the impression, $x_t: \R_{\geq0} \to [0,1]$ denote the allocation function specifying the probability of winning given a bid, and let $p_t : \R_{\geq0} \to [0,1]$ be the payment incurred upon winning the auction.
Formally, the bidder’s optimization problem is given by:
\begin{equation}\label{prob_1}
\begin{array}{ll}
     \underset{b_{t}: t=1, \cdots, T}{\mbox{maximize}} & \sum_{t=1}^T v_t\cdot x_t(b_t)  \\
     \mbox{subject to}&  \sum_{t = 1}^T p_t(b_t) \leq B,
       \\ &\mathsf{RoS} \cdot \sum_{t=1}^T p_t(b_t) \leq \sum_{t=1}^T v_t \cdot x_t(b_t),
\end{array}
\end{equation}
where $B$ is the total budget among the $T$ auctions, and $\mathsf{RoS}>0$ is the target RoS ratio, which requires that the ratio of the advertiser’s total obtained value  $\sum_{t=1}^T v_t \cdot x_t(b_t)$ to the total payment $\sum_{t = 1}^T p_t(b_t)$ be no less than this threshold. Without of loss of generality, we assume $\mathsf{RoS}=1$ through the paper, since we can always rescale the values. 
The problem \eqref{prob_1} is referred to as the offline optimum (or hindsight optimum), as it requires prior knowledge of $v_t, x_t(\cdot)$, and $p_t(\cdot)$ for all $t\in[T]$.

For each time step $t \in [T]$, define $\gamma_t = (z_t,v_t, x_t(\cdot), p_t(\cdot))$ as the auction request, where $z_t\in \mathcal{Z}$ represents the auction's contextual features.
We consider the setting in which only $z_t$ in the auction request is observed by the agent prior to bidding, while $x_t(b_t)$ and $p_t(b_t)$ are revealed at the end of each round,  as the bidding agent observes whether she wins the auction and the corresponding payment.
The value $v_t$ is unknown at the time of the auction and may be revealed with a delay after the auction concludes.
For any pair of feature and value $(z, v)$, we consider the regression model:
\begin{equation}\label{model}
    v =\mu(z) + \epsilon,
\end{equation}
where  $\mu(z)=\E(v \mid z)$ represents the expected effect of the contextual feature on the bidder's value, and $\epsilon=v-\mu(z)$ denotes the intrinsic noise due to the uncertainty of the market. Such a model has also been used in \citet{han2024conformal,banchio2025autobidding}.

Define the sequence of requests as $\vec{\gamma} := (\gamma_1, \dots, \gamma_T)$.
Given any bid vector $\vec{b} = (b_1, \dots, b_T) \in \R_{\geq0}^T$, the bidder’s total reward over the sequence $\vec{\gamma}$ is defined as
$\text{Reward}(\vec{b}, \vec{\gamma}) := \sum_{t=1}^T v_t \cdot x_t(b_t).$

If the $\vec{b}$ is obtained by an algorithm Alg, we denote the reward of Alg for a sequence of requests $\vec{\gamma}$ as $\text{Reward}(\text{Alg}, \vec{\gamma})$.
Let $\vec{b}^{\text{OPT}} = (b_1^{\text{OPT}}, \dots, b_T^{\text{OPT}})$ denote the optimal offline bidding strategy solved by \eqref{prob_1}. The optimal reward is then defined as 
\begin{equation}\label{OPT}
    \text{Reward}(\vec{b}^{\text{OPT}}, \vec{\gamma})=\sum_{t=1}^T v_t \cdot x_t(b_t^{\text{OPT}}).
\end{equation}
By definition, $ \text{Reward}(\vec{b}^{\text{OPT}}, \vec{\gamma}) \geq \text{Reward}(\vec{b}, \vec{\gamma})$ for any bids $\vec{b}$ satisfying the constraints in \eqref{prob_1}, and for any sequence $\vec{\gamma}$.

\subsection{Auto-Bidding with Uncertainty}
Our goal is to ensure that the auto-bidding algorithm achieves a reward comparable to the optimal reward \eqref{OPT} in the case where values $v_1,\dots,, v_T$ are unknown. Previous works \citep{vijayan2025online, feng2018learning} have proposed methods to learn values from online outcome feedback. However, these methods are difficult to implement in practice because they require post-hoc information, such as the realized rewards and the auction outcomes (e.g., the allocation function $x_t(\cdot)$), to infer values and bidding decisions. Such information is often unavailable in a timely manner due to delayed feedback \citep{deng2020calibrating} and the privacy of other bidders’ bidding information.

In practice, the platform learns the value by fitting $\mu(\cdot)$ using the offline data. Formally, assume access to a subset of the bidder’s historical bidding dataset $\mathcal{D} = \{(z^*_j, v^*_j) \mid j = 1, 2, \dots, N\}$, where $z_j^*\in \mathcal{Z}$ denotes the contextual features of the $j$-th auction and $v_j^*\in [0,1]$ represents the bidder's post-hoc true value for the item in that auction. Note that $v_j^*$ is posterior information that may only be observed with a delay after the auction. Such data is readily available in practice, as each bidder typically participates in tens of thousands of auctions per day \citep{deng2020calibrating}. Based on such datasets, one can estimate values via the regression model \eqref{model}. However, these estimates are inevitably subject to uncertainty due to market noise and model fitting \citep{hartline2013mechanism}, which may compromise downstream performance without guarantees. 
To address this limitation, we propose a novel uncertainty-prediction approach that enhances the robustness of auto-bidding while offering formal performance guarantees. We impose two mild assumptions on the data points.

\begin{assumption}[Independent Data]
\label{assum:indep}
The data points  $ \{(z_j^*,v_j^*)_{j=1}^N$, $(z_t, v_t)_{t=1}^T\} $  are independent, though not necessarily identically distributed.
\end{assumption}

\begin{assumption}[Finite Types of Distributions]
\label{assum:tv}
The pair of contextual features and values, $(z, v)$, follows a finite number of distinct distributions, and $\mathcal{D}_{}$ contains data from all such distributions.
\end{assumption}
Assumption \ref{assum:indep} generalizes the i.i.d. assumption commonly used in the literature \citep[e.g.,][]{feng2023online, vijayan2025online}, as we only require the data to be independent, without the need for identical distributions.
Assumption \ref{assum:tv} can be justified in the context of online advertising, where the contextual features of the auction request are categorized into finite fields, such as user group and item category \citep{wei2022posterior, yang2024deep}. 

\section{Main Results}

\subsection{Uncertainty Quantification of the Bidder's Values}

We employ the generalized conformal prediction methods for non-i.i.d. data, as proposed by \citet{barber2023conformal}, to construct a confidence band for each new value. Specifically, we utilize the split conformal prediction method for inference through weighted quantiles. Without loss of generality, let the total number of data points be $N=2n$. We split the data into two equal parts: a \emph{training data} set $\mathcal{D}_{\text{train}}$ and a \emph{calibration data} set $\mathcal{D}_{\text{cali}}$. 
By applying a machine learning method to the training data $\mathcal{D}_{\text{train}}$, one can obtain a pre-trained value prediction model $\hat{\mu}: \mathcal{Z} \to [0,1]$, which is usually provided by the ad platform. 
Using the model, for any $(z,v) \in \mathcal{Z} \times [0,1]$, we define the nonconformity score as:
\begin{equation}\label{hatS}
    \widehat{S}(z,v) := v - \hat{\mu}(z),    
\end{equation}
which measures how unusual the pair $(z,v)$ is relative to the training data. Unlike the widely used nonconformity score $|v - \hat{\mu}(z)|$ in the statistics literature \citep[e.g.,][]{lei2018distribution,barber2023conformal}, we omit the absolute value and instead focus on constructing a one-sided high-confidence interval with an upper bound. This is because our method specifically concerns only the upper bound (see the next section for details). Additionally, without the absolute value, the nonconformity score $\widehat{S}(z,v)$ can take negative values, providing a more precise indication of the difference between the predicted value and the true value.

To simplify the notation, let the calibration dataset be $\mathcal{D}_{\text{cali}}=\{(z_j^*,v_j^*)\mid j=1,2,\dots, n\}$. Given any weights $\{\tilde{\omega}_j^t\}_{j=1}^n$ such that $\tilde{\omega}_j^t\in [0,1]$ and $\sum_{j=1}^n \tilde{\omega}_j^t =1$,  
the nonexchangeable split conformal set for the new value $v_t, t\in [T]$, is given by $\widehat{C}(z_t)
=$  \[\left \{v\in[0,1]:   \widehat{S}(z_t,v)\leq \mathsf{Q}_{1-\beta}\left( \sum_{j=1}^n\tilde{\omega}_j^t\cdot \delta_{\widehat{S}(z_j^*,v_j^*)}+\tilde{\omega}^t_{n+1}\cdot \delta_{+\infty} \right)\right\},\]

\noindent
where $\mathsf{Q}_{\tau}$ denotes the $\tau$-quantile of its argument, and $\delta_{a}$ denotes the points mass at $a$, and $\delta_{+\infty}$ represents a point mass at $+\infty$, i.e., an effectively unbounded value included to ensure proper quantile calibration. When the data points are independent (but not necessarily identically distributed), it holds that  \citep{barber2023conformal}
\begin{equation}\label{eq7}
    \P(v_t\in \widehat{C}(z_t))\geq 1-\beta-2\sum_{j=1}^n \tilde{\omega}_j^t\cdot \mathsf{d}_{\text{TV}}((z_j^*,v_j^*), (z_t,v_t)),
\end{equation}
where $\mathsf{d}_{\text{TV}}$ denotes the total variation distance between distributions of the data.

By substituting \eqref{hatS} into $\widehat{C}(z_{t})$ and defining \[d_t=\mathsf{Q}_{1-\beta}\left( \sum_{j=1}^n\tilde{\omega}_j^t\cdot \delta_{\widehat{S}(z_j^*,v_j^*)}+\tilde{\omega}_{n+1}^t\cdot \delta_{+\infty} \right),\] we can express the prediction interval in a more concise form
$\widehat{C}(z_{t})=[0,\hat{\mu}(z_t)+d_t]:=[0, \hat{v}_t].$
Note that $d_t\in[-1,1]$ and can take both positive and negative values. Intuitively, $d_t$ serves as an adjustment term to the original machine learning prediction $\hat{\mu}(z_t)$ that helps reduce the prediction bias. We define the upper bound of the interval as $ \hat{v}_t:=\hat{\mu}(z_t)+d_t$. The full procedure of getting the prediction interval is summarized in Algorithm \ref{alg:0}.

\begin{algorithm}[!ht]
\caption{Split Conformal Prediction for Non-i.i.d. Data Using Weighted Quantiles.}\label{alg:0}

\begin{algorithmic}[1]
\State \textbf{Input:} Historical data $\mathcal{D} = \{(z^*_j, v^*_j) \mid j = 1, 2, \dots, 2n\}$, miscoverage level $\alpha\in(0,1)$, machine learning algorithm $\mathcal{A}$, total number of auctions $T$,  feature $z_t$, weights $\{\tilde{\omega}_j^t\}_{j=1}^n$.
\State \textbf{Process:} 
    \State Randomly split $\mathcal{D}$ into two disjoint sets: $\mathcal{D}_{\text{train}}$ and $\mathcal{D}_{\text{cali}}=\{(z_j^*,v_j^*)\mid j=1,2,\dots, n\}$.
    \State Fit the value predict function: $\hat{\mu}\leftarrow\mathcal{A}(\mathcal{D}_{\text{train}})$.
    \State Compute the following adjustment term with $\beta=\alpha/T$
    \[d_t=\mathsf{Q}_{1-\beta}\left( \sum_{j=1}^n\tilde{\omega}_j^t\cdot \delta_{\widehat{S}(z_j^*,v_j^*)}+\tilde{\omega}_{n+1}^t\cdot \delta_{+\infty} \right).\]
    \State Calculate the  bidder’s adjusted predicted value $\hat{v}_t=\hat{\mu}(z_t)+d_t$.
\State \Return The uncertainty quantification interval $ \widehat{C}(z_{t})=[0, \hat{v}_t]$ for the unknown true value $v_t$ in the $t$-th auction.
\end{algorithmic} 
\end{algorithm}

Under mild assumptions, the prediction intervals enjoy the coverage guarantee for the new values. 

\begin{theorem}\label{theo_conf}
Under Assumptions \ref{assum:indep} and \ref{assum:tv}, let $\beta = \alpha / T$. By applying all the weights $\{\tilde{\omega}_j^t\}_{j=1}^n$ to the previous data with the same distribution of  $(z_t,v_t)$,  it holds that 
$\P(\forall t, v_t \in \widehat{C}(z_t)) \geq 1 - \alpha.$
\end{theorem}

The proof is provided in Appendix \ref{A1.0}. 
This theorem provides a formal uncertainty quantification for bidder's values based on Algorithm \ref{alg:0}, which does not require the traditional i.i.d. assumption and offers an efficient way to estimate uncertainty prior to new auctions. 
Moreover, we present a straightforward method for selecting the weights that eliminates the total variation distance term in \eqref{eq7} derived by \citet{barber2023conformal}, and we subsequently apply the union bound, which ensures that the coverage guarantee holds uniformly for any $t\in[T]$.

\subsection{Offline Optimum Without Knowing the Values }\label{sec:3.2}

\subsubsection{Reformulation of Offline Optimum Problem.}
Since the value of the bidder for each item is unknown,
using the prediction interval $\widehat{C}(z_{t})$ for each value $v_t$ provided by Algorithm \ref{alg:0}, 
we reformulate \eqref{prob_1} by incorporating the value estimates within their respective prediction intervals, with $\mathsf{RoS} = 1$,
\begin{equation}\label{prob_2}
\begin{array}{ll}
     \underset{\{b_{t}\}_{t=1}^T, \{v_t\}_{t=1}^T}{\mbox{maximize}} & \sum_{t=1}^T v_t\cdot x_t(b_t)  \\
     \mbox{subject to}& \forall t\in[T], v_t\in \widehat{C}(z_{t}),  \\
     &     \sum_{t = 1}^T p_t(b_t) \leq B,
       \\ & \sum_{t=1}^T p_t(b_t) \leq \sum_{t=1}^T v_t \cdot x_t(b_t). 
\end{array}
\end{equation}

Let $\vec{b}^{*}=(b_1^{*},\dots, b_T^{*} )$ and $\vec{v}^{*}=(v_1^{*},\dots, v_T^{*} )$ denote the solution to \eqref{prob_2}. 
Similar to the result in \citet{vijayan2025online} that uses the upper bound of the confidence intervals of the values for bidding, the following proposition shows that in \eqref{prob_2}, the value $v_t^*$ corresponds to the upper bound of the prediction interval, which can be interpreted as the bidder’s adjusted predicted value.

\begin{proposition}\label{prop_1}
For any sequence of requests $\vec{\gamma}$ and for all $t \in [T]$, the optimal value satisfies $v_t^* = \hat{v}_t$.
\end{proposition}

The proof is provided in Appendix \ref{A2.0}.
Compared with \citep{vijayan2025online}, we do not quantify the uncertainty of the allocation and payment functions $x_t(\cdot)$ and $p_t(\cdot)$, since the bidding agent’s strategy typically depends only on the current value and constraints \citep{aggarwal2019autobidding, he2021unified, feng2023online, guo2024generative, aggarwal2025no}. Moreover, the allocation and payment functions cannot be observed when the bids of other participants are not revealed.
By Proposition \ref{prop_1}, \eqref{prob_2} is equivalent to solving the following problem,
\begin{equation}\label{prob_3}
\begin{array}{ll}
    \underset{b_{t}: t=1, \cdots, T}{\mbox{maximize}} & \sum_{t=1}^T \hat{v}_t\cdot x_t(b_t)  \\
     \mbox{subject to}&   \sum_{t = 1}^T p_t(b_t) \leq B,
       \\ & \sum_{t=1}^T p_t(b_t) \leq \sum_{t=1}^T  \hat{v}_t \cdot x_t(b_t). 
\end{array}
\end{equation}

\subsubsection{Reward Guarantee}
Due to the uncertainty of the values, the bids obtained from solving \eqref{prob_3} will incur some reward loss when compared to those from solving \eqref{prob_1}. 
The following theorem provides a reward guarantee for using the adjusted predicted values, i.e., the upper bounds of the prediction intervals, as surrogates for the true values.
We define $d_{\text{max}} = \max_{t \in [T]}  |\hat{v}_t - v_t|$ as the maximum absolute difference between the adjusted predicted value $\hat{v}_t$ and the true value $v_t$ across the $T$ auctions, and $v_{\text{min}} = \min_{t \in [T]} v_t$ as the minimum value among the $T$ auctions.

\begin{theorem}\label{theo_1}
Under Assumptions \ref{assum:indep} and \ref{assum:tv}, for any sequence of requests $\vec{\gamma}$ and any algorithm Alg that outputs bids $\vec{b}=(b_1, \dots, b_T)$, with probability at least $1 - \alpha$, the following holds
\begin{equation*}
\begin{split}
    &\E_{\vec{\gamma}\sim \vec{\mathcal{P}}}\left[\sum_{t=1}^T \hat{v}_t\cdot x_t(b^*_t) -\sum_{t=1}^T \hat{v}_t\cdot x_t(b_t)\right]\\
     \geq&\E_{\vec{\gamma}\sim \vec{\mathcal{P}}}\left[\text{Reward}(\vec{b}^{\text{OPT}}, \vec{\gamma})-\left(1+\frac{d_{\text{max}}}{v_{\text{min}}}\right)\text{Reward}(\text{Alg}, \vec{\gamma})\right],
\end{split}
\end{equation*}
where $\vec{b}^*=(b_1^{*},\dots, b_T^{*} )$ denotes the solution to Problem \eqref{prob_3} and $\vec{b}^{\text{OPT}}$ denotes the solution to Problem \eqref{prob_1}. 
\end{theorem}

The proof is provided in Appendix \ref{A3.0}. 
Throughout this paper, we assume that $v_{\text{min}} > 0$, which is a standard assumption in auction literature \citep{babaioff2017posting}. 
Theorem \ref{theo_1} shows that, for a fixed bidding output, the regret relative to the optimal reward of Problem \eqref{prob_1} with true values is bounded by the regret relative to the offline optimum reward in Problem \eqref{prob_3}, which uses adjusted predicted values as surrogates for the true values.
The result does not depend on the choice of bidding algorithm or auction scheme. In addition, it remains valid regardless of specific constraints imposed by the bidding environment, such as RoS and budget.  Moreover, since our results are derived using values normalized to [0,1], hence the ratio $d_{max}/v_{min}$ in Theorem \ref{theo_1} remains shift-invariant.

Note that Problem \eqref{prob_3} represents the offline optimum with the adjusted predicted values. 
By Theorem \ref{theo_1}, this implies that any online algorithm designed to approximate the offline optimum in Problem \eqref{prob_1} can instead be implemented with the adjusted predicted values as surrogates for the true values.
Specifically,  if the online algorithm achieves a reward close to the optimal reward $\text{Reward}(\vec{b}^{\text{OPT}}, \vec{\gamma})$ defined in \eqref{OPT} when the true values are known, then applying it with the adjusted predicted values and corresponding output bids $\vec{b}=(b_1, \dots, b_T)$ guarantees that $\sum_{t=1}^T \hat{v}_t\cdot x_t(b_t)$ is also close to $\sum_{t=1}^T \hat{v}_t\cdot x_t(b^*_t)$, which represents the optimal reward using the adjusted values. Thus, by Theorem \ref{theo_1}, the reward under the bids $\vec{b}$ is comparable to the optimal reward (see Section \ref{3.3} for a detailed analysis using a specific online algorithm).
By contrast, directly relying on the machine-learning prediction $\hat{\mu}(z)$ makes it difficult to provide such guarantees since market noise $\epsilon=v - \mu(z)$ defined in \eqref{model} introduces irreducible uncertainty that cannot be learned using the feature $z$. Our adjustment procedure overcomes this limitation and establishes a provable reward guarantee.

\subsubsection{RoS Guarentee}
In addition to the reward guarantee, we present the following result, which establishes that, for any bidding strategy, the RoS under the adjusted predicted values and the RoS under the true values bound each other.  
Given the bids $\vec{b}$ for a sequence of requests $\vec{\gamma}$, define
\begin{equation*}
    \text{RoS}(\vec{b}, \vec{\gamma})=\frac{\sum_{t=1}^T v_t \cdot x_t(b_t)}{\ \sum_{t=1}^T p_t(b_t)}  \text{\quad  and\quad } \widehat{\text{RoS}}(\vec{b}, \vec{\gamma})=\frac{\sum_{t=1}^T \hat{v}_t \cdot x_t(b_t)}{\ \sum_{t=1}^T p_t(b_t)},
\end{equation*}
where $\text{RoS}(\vec{b}, \vec{\gamma})$ denotes the RoS under the true values, while $\widehat{\text{RoS}}(\vec{b}, \vec{\gamma})$ represents the RoS under the adjusted predicted values.

\begin{theorem}\label{thoe:3}
   Given any bid vector $\vec{b} = (b_1, \dots, b_T) \in \R_{\geq0}^T$ and any sequence of requests $\vec{\gamma}$, the following inequality holds
\begin{equation*}
\left (1-\frac{d_{\text{max}}}{v_{\text{min}}}\right)\text{RoS}(\vec{b}, \vec{\gamma})\leq \widehat{\text{RoS}}(\vec{b}, \vec{\gamma})\leq \left (1+\frac{d_{\text{max}}}{v_{\text{min}}}\right)\text{RoS}(\vec{b}, \vec{\gamma}).  
\end{equation*}
\end{theorem}

The proof is provided in Appendix \ref{A3.1}.
This theorem \ref{thoe:3} establishes that if an online algorithm for Problem \eqref{prob_3} outputs bids $\vec{b}=(b_1, \dots, b_T)$ achieving a favorable $\widehat{\text{RoS}}(\vec{b}, \vec{\gamma})$ under the adjusted predicted values, then the corresponding RoS under the true values, $\text{RoS}(\vec{b}, \vec{\gamma})$, is also guaranteed to be well bounded.  Specifically, it bounds the ratio of $\text{RoS}(\vec{b}, \vec{\gamma})$ to $\widehat{\text{RoS}}(\vec{b}, \vec{\gamma})$, with the guarantee arising from the small discrepancy between the true values and the adjusted predicted values, and it holds for any allocation and payment scheme. With this guarantee, any algorithm that achieves good RoS for Problem \eqref{prob_1} can be applied to Problem \eqref{prob_3}, with the adjusted predicted values serving as surrogates for the true values, thus providing a robust theoretical guarantee even when the true values are unknown.

\subsubsection{Optimizing Performance}
To improve the reward and RoS guarantees, we aim to reduce the value of $d_{\text{max}}$, as suggested by Theorem \ref{theo_1} and Theorem \ref{thoe:3}. 
We present the following proposition, which shows that $d_{\text{max}}$ can be upper bounded by the machine learning prediction error and the value approximation error.

\begin{proposition}\label{prop2}
The following upper bound holds for $d_{\text{max}}$
\begin{equation*}
  d_{\text{max}} \leq 2 \cdot \left( \sup_{z} \left| \hat{\mu}(z) - \mu(z) \right| + \sup_{(z,v)} \left| v - \mu(z) \right| \right),  
\end{equation*}
where $\sup_{z} \left| \hat{\mu}(z) - \mu(z) \right|$ represents the prediction error of the machine learning algorithm, and $\sup_{(z,v)} \left| v - \mu(z) \right|$ denotes the value approximation error. The supremum is taken over all pairs of feature and value $(z, v)$ in the space $\mathcal{Z} \times [0,1]$.
\end{proposition}

The proof is provided in Appendix \ref{Ap.2}. 
Therefore, to achieve improved performance, it is beneficial to employ more accurate machine learning algorithms to better fit the model to the data, thereby reducing prediction error. Additionally, incorporating richer features, such as post-click signals \citep{deng2020calibrating}, can help reduce approximation error. The problem of accurately predicting values for online advertising has been actively studied in recent years \citep{wei2022posterior, yang2024deep}. Our method is compatible with these approaches and further enables uncertainty quantification of the values, while providing guarantees on reward and RoS for downstream performance.

\subsection{Online Bidding Without Knowing the Values} \label{3.3}

In practice, the bid at time $t$ is based only on the auction outcomes from the previous $t-1$ steps and the contextual features available at time $t$ \citep{banchio2025autobidding}. Consequently, the offline optimum cannot be directly computed, as it requires knowledge of all auction outcomes and item values across the $T$ auctions. Therefore, an online algorithm is needed to use the available information for approximating the offline optimum. Prior work has focused on online algorithms that approximate Problem \eqref{prob_1} under the assumption that item values are known, an assumption that does not hold in practice \citep{deng2020calibrating, wei2022posterior, yang2024deep}. In this section, we extend the method from Section \ref{sec:3.2} to the online setting where the true values are unknown.

We present an adjusted online algorithm for truthful auctions, building directly on the online bidding framework of \citet{feng2023online}, which itself applies the dual mirror descent method of \citet{balseiro2020dual}. 
The truthful auctions, such as the second-price auction, are widely used in practice \citep[][]{mohri2016learning, ostrovsky2023reserve}.
While the algorithm of \citet{feng2023online} assumes access to true bidder values and i.i.d. auction requests,  we extend their approach to a more practical setting where the auto-bidder does not observe the true values prior to bidding and auction requests are not necessarily identically distributed. Specifically, our adaptation relies on adjusted predicted values, with the auto-bidder only observing contextual features before each auction. The adjusted bidding algorithm is presented in Algorithm \ref{alg:1}.

\begin{algorithm}[!ht]
\caption{Online Bidding Algorithm for Non-i.i.d.\ Inputs Without Knowing the Values}\label{alg:1}

\begin{algorithmic}[1]
\State \textbf{Input:} Total number of auctions $T$, feature $z_t$, total budget $B>0$, and $\mathsf{RoS}=1$.
\State \textbf{Initialize:} Initial dual variable $\lambda_1=1$, $\mu_1=0$, total initial budget $B_1 := B$,  dual mirror descent step size $\phi = \frac{1}{\sqrt{T}}$ and $\eta = \frac{1}{(1+(B/T)^2)\sqrt{T}}$.

\For{$t=1,2,\cdots, T$}
    \State Observe the feature $z_t$, 
    compute the $\hat{v}_t$ using Algorithm \ref{alg:0}.
    
    \State Set the bid $b_t = 
\begin{cases}
\frac{1+  \lambda_{t}}{\mu_t + \lambda_{t}}\cdot \hat{v}_t, & B_t \geq 1, \\
0, & \text{otherwise}.
\end{cases}$

    \State Compute  $g_t(b_t) = \hat{v}_t \cdot x_t({b}_t) - p_t(b_t)$.

    \State Update the dual variable of the RoS constraint as $\lambda_{t+1} = \lambda_t \exp\left[-\phi\cdot g_t(b_t)\right]$.  
    
    \State Compute $g_t^{\prime}(b_t) = B/T- p_t(b_t)$.

    \State Update the dual variable of the budget constraint as $    \mu_{t+1} := \mathrm{Proj}_{\mu \geq 0}(\mu_t - \eta \cdot g_t^{\prime}(b_t))$.  
    
    \State Update the leftover budget $B_{t+1} = B_t - p_t(b_t)$. 
\EndFor
\State \Return The sequence $\{b_t\}_{t=1}^T$ of bids. 
\end{algorithmic} 
\end{algorithm}

We denote $\vec{\mathcal{P}}$ the joint distribution over the input sequence $\vec{\gamma}$ over $T$ auctions, and $\mathcal{P}_t$ denotes the distribution for the request $\gamma_t$. Let $\text{MD}(\vec{\mathcal{P}})=\sum_{t=1}^T ||\mathcal{P}_t-\frac{1}{T}\sum_{s=1}^T\mathcal{P}_s||_{\text{TV}}$ represent the mean deviation of the vector of independent distributions $\vec{\mathcal{P}}$ from the average distribution, measured in total variation distance. 
Building on the results of Section \ref{sec:3.2}, we now analyze the reward and constraint violations of Algorithm \ref{alg:1}.

\begin{theorem}\label{theo_3}
Under Assumptions \ref{assum:indep} and \ref{assum:tv}, and assuming that ${B}/{T} = O(1)$ and $\text{MD}(\vec{\mathcal{P}}) = O(1)$, for inputs drawn from any joint distribution $\vec{\mathcal{P}}$ over $T$ truthful auctions, with probability at least $1 - \alpha$, the following holds
    \[E_{\vec{\gamma}\sim \vec{\mathcal{P}}}\left[ \frac{1}{1+\xi}\text{Reward}(\vec{b}^{\text{OPT}}, \vec{\gamma})-\text{Reward}(\text{Alg}, \vec{\gamma})\right]\leq  O(\sqrt{T}),\]
where $\xi = {d_{\text{max}}}/v_{\text{min}}$.
Moreover, $\sum_{t=1}^T p_t(b_t)\leq B$ that implies no violation of the budget constraint. For the RoS constraint,  the following holds for any sequence of requests $\vec{\gamma}$
\begin{equation}\label{e9}
    \sum_{t=1}^T p_t(b_t)-\sum_{t=1}^T v_t \cdot x_t(b_t)\cdot\left (1+\frac{d_{\text{max}}}{v_{\text{min}}}\right)\leq O(\sqrt{T}logT).
\end{equation}
\end{theorem}

The proof is provided in Appendix \ref{A4.0}. The assumption ${B}/{T} = O(1)$ is standard in the auto-bidding literature  \citep{feng2023online}, where it is commonly assumed that ${B}/{T} = \rho$ for some constant $\rho > 0$.  Similarly, the assumption $\text{MD}(\vec{\mathcal{P}}) = O(1)$ is also employed in \citet{balseiro2023best}, where it is noted that this measure of non-stationarity in the distributions is closely related to the Kolmogorov metric, which has been used in the study of posted pricing \citep{dutting2019posted}.

Theorem \ref{theo_3} shows that Algorithm \ref{alg:1} achieves near-optimal bounds, attaining a competitive ratio against the optimal reward and ensuring near-optimal bounds on RoS violation
in the non-i.i.d. setting without knowing the values. 
As noted by \citet{balseiro2023best}, in the non-i.i.d. setting, analyzing the competitive ratio is more meaningful than considering regret. Thus, our reward bound establishes a new competitive ratio in this setting. The competitive ratio in our result is $1 + \xi$, where $\xi$ characterizes the uncertainty in the bidders’ values, as derived from Theorem \ref{theo_1}. This ratio depends on the auction requests, consistent with the observation of \citet{aggarwal2024auto} that fixed competitive ratios are generally unattainable.

Additionally, with respect to the RoS violation, compared to the results in \citet{feng2023online}, the inequality \eqref{e9} introduces an additional term $d_{\text{max}}/v_{\text{min}}$ due to the uncertainty in the predicted values, which stems from Theorem \ref{thoe:3}. However, the convergence rate remains near-optimal. 

\section{Numerical Experiments}

We conduct experiments on datasets where each bidder aims to maximize the number of conversions, subject to budget and tCPA constraints. In this setting, the value of a bid is determined by the ad’s CVR, while the RoS constraint is captured by the tCPA, set as $\mathsf{tCPA}$ by the advertiser. At time $t$, let $\mathsf{CVR}_t$ denote the actual CVR, which is unknown all the time, $\mathsf{pCVR}_t$ the predicted CVR, which can be obtained before the auction, and $\mathsf{CVR}_t^{\text{true}}$ the post-hoc true CVR, which can be obtained after the auction. Following Problem~\eqref{prob_1} with $\mathsf{RoS}=1$, we rescale the value as $v_t = \mathsf{CVR}_t \cdot \mathsf{tCPA}$, so that $\mathsf{tCPA}$ can be normalized to 1, and the uncertainty in the value arises from the prediction of the conversion rate, i.e., $\mathsf{pCVR}_t$.

\noindent
\textbf{Compared Methods.}
We compare four different bidding methods. The first three are based on the no-regret algorithm proposed by \citet{feng2023online} that uses online mirror descent:
\begin{itemize}[leftmargin=*]
\item Online Mirror Descent with Adjusted Values (\textbf{Adjust}): Directly applies Algorithm \ref{alg:1} using the adjusted predicted value $\hat{v}_t=(\mathsf{pCVR}_t + d_t) \cdot \mathsf{tCPA}$.
\item Online Mirror Descent with Predicted Values (\textbf{Pred}): Applies Algorithm \ref{alg:1} using the predicted values $\hat{\mu}(z_t) = \mathsf{pCVR}_t \cdot \mathsf{tCPA}$ in place of the adjusted value $\hat{v}_t$.
\item Online Mirror Descent with Post-Hoc True Values (\textbf{True}): Applies Algorithm \ref{alg:1} using the post-hoc true values $v_t = \mathsf{CVR}_t^{\text{true}} \cdot \mathsf{tCPA}$ instead of the adjusted value $\hat{v}_t$.
\item Upper Confidence Bound-style Algorithm (\textbf{UCB}): A state-of-the-art online bidding algorithm proposed by \citet{vijayan2025online} for settings with budget and RoS constraints, without requiring knowledge of the true values. This method requires knowledge of whether a conversion occurs after the auction, which we simulate as $\text{conversion} \sim \text{Binomial}(\mathsf{CVR}_t^{\text{true}})$. 
The bidding set of a bidder is a uniformly spaced grid between 0 and the maximum bid of the bidder in the dataset, consisting of 50 points.
\end{itemize}

\noindent
\textbf{Evaluation Metrics.}
Performance of all methods is evaluated using cumulative expected conversions, adjusted by a penalty term when the realized CPA exceeds a predefined target. The final metric is defined as
\begin{equation}\label{score}
\text{Score} = \mathbb{P} \sum_{t=1}^T \mathsf{CVR}_t^{\text{true}} \cdot x_t(b_t), \quad
\mathbb{P} = \min\left\{\left(\frac{C}{\mathsf{tCPA}}\right)^{\zeta}, 1\right\},
\end{equation}
where $\mathbb{P}$ denotes the penalty function for exceeding the target CPA, $C$ is the realized CPA, and $\zeta$ is a hyperparameter set to 2 to control the magnitude of the penalty.

Different bidders may achieve distinct total rewards and, consequently, scores of varying magnitudes.
To enable a fair comparison across different methods applied to all bidders, we normalize the score using the \textbf{True} algorithm as a baseline and define the performance ratio:
\begin{equation}\label{ratio_alg}
\text{Ratio(Alg)} = \frac{\E_{\vec{\gamma} \sim \vec{\mathcal{P}}}[\text{Score(Alg)}]}{\E_{\vec{\gamma} \sim \vec{\mathcal{P}}}[\text{Score(\textbf{True})}]},
\end{equation}
where $\text{Score(Alg)}$, defined in \eqref{score}, denotes the score achieved by the algorithm {Alg} introduced in the compared methods.
The \textbf{True} algorithm relies on post-hoc true values and therefore represents the optimal revenue in the online setting.

\subsection{Simulation Study}\label{sec:simu}
\subsubsection{Datasets}

We use public datasets from the Alibaba for the simulated bidding environment \citep{su2024auctionnet}, which covers the first eight delivery periods (7–15). Each period can be regarded as a separate time horizon, such as a new day, with about 500,000 ad impression opportunities and 48 competing bidding agents. For each ad opportunity, the dataset provides the predicted conversion rate $\mathsf{pCVR}_t$ and its uncertainty $\sigma_t$, with $\mathsf{CVR}_t \sim \mathcal{N}(\mathsf{pCVR}_t, \sigma_t^2)$ \citep{su2024auctionnet}.  
It also specifies each agent’s budget, $\mathsf{tCPA}$, and bid prices. We consider a single-slot second-price auction, in which the highest bidder wins the slot and, once winning, is charged the second-highest bid.

\subsubsection{Implementation Details}
We use period 7 for training and periods 8–14 for testing. 
Since the dataset already contains the predicted conversion rate $\mathsf{pCVR}_t$, obtained from machine learning models trained with feature data, we use all the training dataset as the calibration set $\mathcal{D}_{\text{cali}}$.
Following \citet{deng2020calibrating, wei2022posterior, yang2024deep}, the post-hoc true $\mathsf{CVR}_t^{\text{true}}$ is estimated across different fields, each corresponding to a specific classified category of the predicted CVR. Specifically, following the method in \citet{deng2020calibrating}, we sort the predicted conversion rates in ascending order and partition them into 100 bins of equal size, treating each bin as a distinct field. 
Within each bin, we simulate the $\mathsf{CVR}_t$ from $ \mathcal{N}(\mathsf{pCVR}_t, \sigma_t^2)$ and compute the average CVR in the bin as the post-hoc true CVR, i.e.,  $\mathsf{CVR}_t^{\text{true}}$.
The use of the average value as a baseline is consistent with prior work such as the UCB-based algorithm \citep{vijayan2025online}. We assume that values within each bin follow the same distribution and, for each new auction, evenly distribute the weights across the historical data belonging to the corresponding bin to construct prediction intervals using Algorithm \ref{alg:0}, with the miscoverage level set to $\alpha=0.1$. For the test sets, we apply the same binning scheme as in the training set to determine the $\mathsf{CVR}_t^{\text{true}}$.

\begin{figure}[t]
  \centering
      \begin{subfigure}{0.2\textwidth}
        \includegraphics[width=0.9\textwidth]{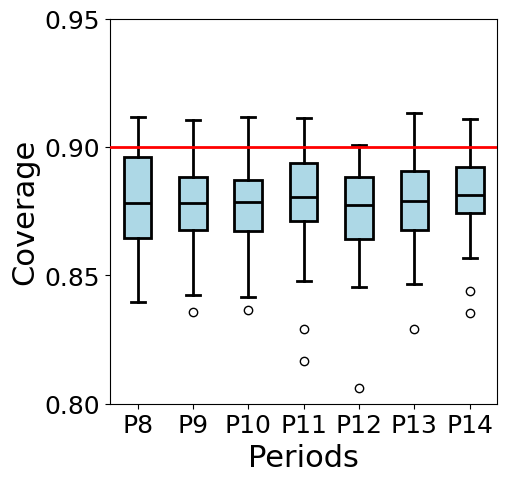} 
        \centering
        \caption{Results of Section \ref{sec:simu}}
       \label{fig:figure1a}
    \end{subfigure}
     \hspace{0.02\textwidth} 
   \begin{subfigure}{0.2\textwidth}
    \includegraphics[width=0.9\textwidth]{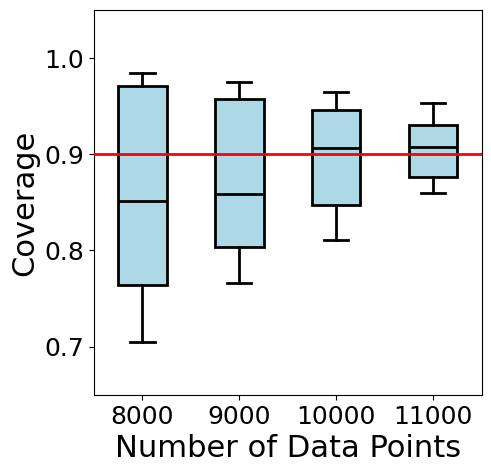} 
        \centering
        \caption{Results of Section \ref{sec:lable}}
    \label{fig:figure1b}
   \end{subfigure}
    \caption{ The average coverage probability of the prediction interval for the post-hoc true values. The red line denotes the target level of $1-\alpha=0.9$.}
\end{figure}

\begin{table}[t]
\centering
\caption{Performance comparison across algorithms over different periods. The best performance of each period is \textbf{bold}.}
\label{tab:per_ali}

\resizebox{\columnwidth}{!}{%
\begin{tabular}{ccccccccc}
\toprule
\( \textbf{Method} \) & P8 & P9 & P10 & P11 & P12 & P13 & P14 & AVG $\pm$ STD \\  
\midrule
UCB    & 0.188 & 0.188 & 0.202 & 0.202 & 0.182 & 0.160 & 0.157 & 0.186 $\pm$ 0.021 \\  
Pred   & 0.980 & 0.979 & 0.985 & 0.992 & 0.980 & 0.986 & 0.996 & 0.986 $\pm$ 0.006 \\ 
Adjust & \textbf{0.987} & \textbf{0.985} & \textbf{0.986} & \textbf{0.999} & \textbf{0.984} & \textbf{0.993} & \textbf{1.000} & \textbf{0.991} $\pm$ \textbf{0.006} \\
\bottomrule
\end{tabular}%
}
\end{table}

\subsubsection{Performance Evaluation}

Figure~\ref{fig:figure1a} presents standard quartile boxplots of the average coverage probabilities of the prediction interval for the post-hoc true values across bidding agents, for each of the seven test periods (periods 8–14), where each agent’s average is computed over all auctions within the period. 
The coverage probabilities remain generally high, exceeding 0.85 across all periods, demonstrating the robustness of our method. Since the distribution of CVR may differ between the training and test data within the same bins, there is a slight deviation from the target level of 0.9.

Table \ref{tab:per_ali} reports the performance ratio \eqref{ratio_alg} of different algorithms across the seven test periods, where the ratio is averaged over all bidding agents. Among the three methods, using the adjusted predicted value consistently achieves the best performance, demonstrating the reward guarantee of our uncertainty quantification approach. 
Note that it requires only a small additional computation to achieve the improvement (see Appendix \ref{add_simu}).
The difference between Pred and Adjust is small because 
$\mathsf{pCVR}_t$ is assumed to be an unbiased prediction of $\mathsf{CVR}_t$, i.e., $\mathsf{CVR}_t=\mathsf{pCVR}_t+\mathcal{N}(0, \sigma_t^2)$, and the variance of the prediction in the dataset is so low that the simulated $\mathsf{CVR}_t^{\text{true}}\approx \mathsf{pCVR}_t$. By contrast, in the next section we demonstrate a distinct difference using real-world data. 
The reward of UCB is lower, as the distribution of the data differs and the conversion rate in the dataset is small, making it difficult to estimate the average CVR without machine learning predictions.

\subsection{Real World Application}\label{sec:lable}

\subsubsection{Datasets}
We use real-world datasets obtained from the bidding feedback logs of an e-commerce platform, which cover more than 80 campaigns and over 23 million ad impression opportunities collected between July 28th and July 29th, 2025, with more than 11 million ad opportunities recorded each day.
For each bid request, the log records comprehensive information, including auction logs, advertiser details (e.g., budget, and tCPA), and user data (e.g., pCVR,  CVR feedback, and pCTR).
We consider a single-slot setting in which the winner is the bidder with the highest estimated cost per mille (eCPM), computed as $b_t \cdot \mathsf{pCTR}_t \cdot 1000$, where $\mathsf{pCTR}_t$ denotes the predicted click-through rate (pCTR) of the bidder at time $t$. The accuracy of $\mathsf{pCTR}_t$ does not affect our analysis, as the same prediction is used across all bidding strategies. The allocation rule $x_t(b_t)$ takes the value 1 if and only if the bidder wins the slot and the user clicks on the ad, and 0 otherwise. The payment rule follows Myerson’s envelope formula, $p_t(b) = b \cdot x_t(b) - \int_0^b x_t(z) \, dz$, 
which guarantees the truthfulness of the mechanism.

\begin{table}[t]
\centering
\setlength\tabcolsep{3pt} %
\fontsize{8}{9}\selectfont %
\caption{Performance comparison of algorithms across different numbers of historical data $n$ and new auctions $T$. The best performance in each setting is highlighted in \textbf{bold}.}
\resizebox{\columnwidth}{!}{ %
\begin{tabular}{cccccccccc}
\toprule 
& \multicolumn{3}{c}{\( n = 6000 \)} & \multicolumn{3}{c}{\( n = 8000 \)} & \multicolumn{3}{c}{\( n = 10000 \)} \\ 
\cmidrule(l{0.5em}r{0.5em}){2-4} \cmidrule(l{0.5em}r{0.5em}){5-7} \cmidrule(l{0.5em}r{0.5em}){8-10}
\( T \) & UCB & Pred & Adjust & UCB & Pred & Adjust & UCB & Pred & Adjust  \\  \midrule
1500  & 0.875 & 0.944 & \textbf{0.984} & 0.886 & 0.945 & \textbf{0.989} & 0.867 & 0.945 & \textbf{0.986} \\  
1900  & 0.849 & 0.939 & \textbf{0.979} & 0.863 & 0.939 & \textbf{0.982} & 0.864 & 0.940 & \textbf{0.983}\\ 
2300  & 0.876 & 0.939 & \textbf{0.981} & 0.856 & 0.938 & \textbf{0.980} & 0.864 & 0.937 & \textbf{0.982}\\
2700  & 0.865 & 0.936 & \textbf{0.977} & 0.871 & 0.938 & \textbf{0.979}& 0.874 & 0.938 & \textbf{0.981}\\
\bottomrule 
\end{tabular}%
}
\label{tab:per_jd} 
\end{table}

\subsubsection{Implementation Details}
We use data from July 28th for training and data from July 29th for testing.
In the absence of ground-truth conversion rates, we follow the approach of \citet{deng2020calibrating} to construct realized post-hoc values: we partition the predicted conversion rates into 100 bins based on their magnitudes, and for each bin compute the average observed conversion, which serves as its post-hoc true value $\mathsf{CVR}_t^{\text{true}}$. 
We treat the values within each bin as having the same distribution, and for each new auction, we evenly distribute all weights across the historical data corresponding to the auction’s value bin to construct the prediction intervals using Algorithm \ref{alg:0}, with the miscoverage level set to $\alpha=0.1$.
On the test set, we used the same bins as in the training set to determine the true values.
In each test experiment, we randomly sample $T$ auctions as $\vec{\gamma}$ from the test set and simulate post-hoc clicks according to $\text{click} \sim \text{Binomial}(\mathsf{pCTR}_t)$. The allocation indicator $x_t(b_t)$ equals 1 if and only if $\text{click} = 1$ and the bidder wins.

\subsubsection{Performance Evaluation}
Figure \ref{fig:figure1b} presents standard quartile boxplots of the average coverage probabilities of the prediction intervals for the post-hoc true values, computed across all bidding agents for different numbers of calibration data points $n$ on all test data. As $n$ increases, the coverage probabilities remain consistently high and converge to the target level, demonstrating the efficiency of the prediction intervals.%

Table \ref{tab:per_jd} presents the performance ratio \eqref{ratio_alg} of different algorithms for $T \in \{1500, 1900,$ $ 2300, 2700\}$ and $n \in \{6000, 8000, 10000\}$, where the ratio for each bidder is averaged over 50 experiments in each setting, and the average performance across all bidders is reported. In all settings, using the adjusted value consistently yields the best performance, demonstrating the efficiency of our uncertainty quantification approach. As $n$ increases, the performance improves, as the conformal prediction interval achieves higher coverage probability, as shown in Figure \ref{fig:figure1b}. Although performance decreases with increasing $T$ due to higher regret, our method still guarantees low regret in this non-i.i.d. and unknown-value setting, maintaining high performance. Compared with Table \ref{tab:per_ali}, UCB performs better in the real-world data setting, as advertisers in the real-world dataset typically have higher CPAs, and the conversion process is easier, allowing UCB to more effectively learn the value from the auction outcomes.

\begin{table}[t]
\centering
\setlength\tabcolsep{3pt} 
\fontsize{8}{9}\selectfont
\caption{Average computation time (in seconds) of different algorithms across different numbers of historical data $n$ and new auctions $T$.}
\resizebox{\columnwidth}{!}{ %
\begin{tabular}{cccccccccc}
\toprule 
& \multicolumn{3}{c}{\( n = 6000 \)} & \multicolumn{3}{c}{\( n = 8000 \)} & \multicolumn{3}{c}{\( n = 10000 \)} \\ 
\cmidrule(l{0.5em}r{0.5em}){2-4} \cmidrule(l{0.5em}r{0.5em}){5-7} \cmidrule(l{0.5em}r{0.5em}){8-10}
\( T \) & UCB & Pred & Adjust & UCB & Pred & Adjust & UCB & Pred & Adjust  \\  \midrule
1500  & 50.269 & 0.005 & 0.016 & 45.484 & 0.006 & 0.016 & 43.743 & 0.005 & 0.014 \\  
1900  & 67.789 & 0.007 & 0.019 & 43.183 & 0.007 & 0.021 & 65.382 & 0.007 & 0.017\\ 
2300  & 68.828 & 0.008 & 0.022 & 51.018 & 0.010 & 0.031 & 61.361 & 0.008 & 0.020\\
2700  & 60.789 & 0.010 & 0.025 & 58.591 & 0.012 & 0.037 & 61.431 & 0.009 & 0.024\\
\bottomrule 
\end{tabular}%
}
\label{tab:time_jd} 
\end{table}

Table \ref{tab:time_jd} reports the average computation time (in seconds) of different algorithms across varying numbers of historical data $n$ and new auctions $T$, similar to Table \ref{tab:per_jd}. The time cost is reported as the wall-clock time on an Intel Xeon 6967P-C processor with 32 cores enabled for parallel computing, 180 GB of memory, and no GPU acceleration. The results show that the computational cost of UCB is significantly higher, as it requires solving a linear program for each auction. In contrast, our method leverages machine learning predictions to provide uncertainty quantification of the values, offering high performance with minimal additional computational resources, making it more cost-effective for online advertising.

\section{Conclusion}
This paper studies the online advertising setting where true ad impression opportunity values are unknown. We propose a method to quantify the uncertainty of machine learning predictions using historical bidding data and contextual features. Our approach extends beyond the i.i.d. assumption, is compatible with industry systems, and provides performance guarantees for existing algorithms without perfect knowledge of the true values. We apply it to auto-bidding algorithms under non-i.i.d. requests and validate its effectiveness on simulated Alibaba data and a real-world e-commerce dataset, demonstrating consistent performance gains with computational efficiency.
Future directions include exploring the use of value prediction intervals in broader applications, such as designing auction mechanisms that avoid value distribution estimation while balancing long-term rewards with short-term rewards.

\begin{acks}
We would like to thank the area chair and four anonymous referees for constructive suggestions that improve the paper. Dai's research was supported in part by a Merck Research
Award and a Hellman Fellowship Award.
\end{acks}
  


\newpage

\bibliographystyle{ACM-Reference-Format}
\balance
\bibliography{autobid}


\appendix

\section{Proofs of Main Results}\label{app:proofs}
\subsection{Proof of Theorem \ref{theo_conf}}\label{A1.0}
\begin{proof}~
\noindent
Let $E_t$ denote the event that $v_t\notin\widehat{C}(z_{t})$. Then, we have
    \begin{equation}\label{eq9}
    \begin{split}
    \P(\forall t, v_t\in \widehat{C}(z_{t}))&=1-\P(\exists t, v_t\notin\widehat{C}(z_{t}))=1-\P(\cup_{t=1}^T E_i).
    \end{split}
    \end{equation}
    
Using the union bound, we obtain
\begin{equation}\label{eq10}
    \P(\cup_{t=1}^T E_t)\leq \sum_{t=1}^T \P(E_t).
\end{equation}

Under Assumptions \ref{assum:indep}, and by \eqref{eq7}, we can bound $\P(E_t)$ as follows
\begin{equation}\label{eq11}
\begin{split}
    \P(E_t)\leq \beta+2\sum_{j=1}^n \tilde{\omega}_j^t\cdot \mathsf{d}_{\text{TV}}((z_j^*,v_j^*), (z_t,v_t)).
\end{split}
\end{equation}

Under Assumption \ref{assum:tv}, for each $t\in[T]$, there exist some data points in $(z_j^*,v_j^*)_{j=1}^n$ such that the total variation distance between the distribution of these data and the distribution of $(z_t, v_t) $ is zero.
For each $t\in[T]$,  we set the weights in the set \[\{\tilde{\omega}_j^t\mid 1\leq j\leq n,  \mathsf{d}_{\text{TV}}\left( (z_j^*, v_j^*), (z_t, v_t) \right)=0\}\] to be non-negative, while all other weights are set to zero. It then follows that
\begin{equation}\label{eq13}
2\sum_{j=1}^n \tilde{\omega}_j^t\cdot  \mathsf{d}_{\text{TV}}\left( (z_j^*, v_j^*), (z_t, v_t) \right)=0.  
\end{equation}

Combining \eqref{eq9}, \eqref{eq10}, \eqref{eq11}, and \eqref{eq13}, we obtain \[ \P(\forall t, v_t\in \widehat{C}(z_{t}))\geq 1-T\cdot\beta =1-\alpha.\]

This completes the proof of Theorem \ref{theo_conf}.
\end{proof}

\subsection{Proof of Proposition \ref{prop_1}}\label{A2.0}
\begin{proof}~
\noindent
   We first observe that $(\{b_t^{*}\}_{t=1}^T, \{\hat{v}_t\}_{t=1}^T)$ satisfies all the constraints in Problem \eqref{prob_2}. Since $(\{b_t^{*}\}_{t=1}^T, \{{v}_t^*\}_{t=1}^T)$ is the optimal solutions, it follows that
    \begin{equation}\label{17}
        \sum_{t=1}^T v_t^*\cdot x_t(b_t^*)\geq \sum_{t=1}^T \hat{v}_t\cdot x_t(b_t^*).
    \end{equation}
     Next, by the constraint that for all $t \in [T]$, $v_t \in \widehat{C}(z_t)$, we know that $v_t^* \leq \hat{v}_t$ for each $t$. Thus, from \eqref{17}, we conclude that $v_t^* = \hat{v}_t$ for all $t \in [T]$.
    This completes the proof of Proposition \ref{prop_1}.
\end{proof}

\subsection{Proof of Theorem \ref{theo_1}}\label{A3.0}
\begin{proof}~
\noindent With probability at least $1 - \alpha$, for all $t \in [T]$, we have $v_t \in \widehat{C}(z_t)$.
   Therefore, $(\{b_{t}^{\text{OPT}}\}_{t=1}^T, \{v_t\}_{t=1}^T)$ is feasible for Problem \eqref{prob_2}. Since $(\{b_{t}^{*}\}_{t=1}^T, \{v_t^*\}_{t=1}^T)$ represents the optimal solution of Problem \eqref{prob_2}, it follows that
   \begin{equation}\label{new_17}
       \sum_{t=1}^T v_t^*\cdot x_t(b_t^*)\geq \sum_{t=1}^T v_t \cdot x_t(b_t^{\text{OPT}}).
   \end{equation}

   From Proposition \ref{prop_1} and \eqref{new_17}, we have that
\begin{equation}\label{new_36}
\begin{split}
        \E_{\vec{\gamma}\sim \vec{\mathcal{P}}}\left[\sum_{t=1}^T \hat{v}_t\cdot x_t(b^*_t)\right]&\geq \E_{\vec{\gamma}\sim \vec{\mathcal{P}}}\left[\text{Reward}(\vec{b}^{\text{OPT}}, \vec{\gamma})\right].
\end{split}
\end{equation}

From the definitions of $d_{\text{max}}$, we obtain the following inequality
\begin{equation}\label{app_20}
\begin{split}
\frac{ \sum_{t=1}^T \hat{v}_t \cdot x_t(b_t)}{\sum_{t=1}^T {v}_t \cdot x_t(b_t) } \leq \frac{ \sum_{t=1}^T ({v}_t+d_{\text{max}}) \cdot x_t(b_t)}{\sum_{t=1}^T {v}_t \cdot x_t(b_t) }.  
\end{split}
\end{equation}
Furthermore, from the definitions of $ {v}_{\text{min}} $, we have
\begin{equation}\label{app_21}
\begin{split}
  \frac{ \sum_{t=1}^T ({v}_t+d_{\text{max}}) \cdot x_t(b_t)}{\sum_{t=1}^T {v}_t \cdot x_t(b_t) } 
      \leq 1+\frac{ d_{\text{max}} \cdot\sum_{t=1}^T  x_t(b_t)}{ {v}_{\text{min}} \cdot\sum_{t=1}^T x_t(b_t) }=1+\frac{d_{\text{max}}}{v_{\text{min}}}.
\end{split}
\end{equation}

Combining \eqref{app_20} and \eqref{app_21}, we have that 
\begin{equation}\label{app_22}
    \frac{ \sum_{t=1}^T \hat{v}_t \cdot x_t(b_t)}{\sum_{t=1}^T {v}_t \cdot x_t(b_t) } \leq1+\frac{d_{\text{max}}}{v_{\text{min}}}.
\end{equation}

By \eqref{app_22}, it holds that
\begin{equation}\label{new_37}
\begin{split}
         \E_{\vec{\gamma}\sim \vec{\mathcal{P}}}\left[\sum_{t=1}^T \hat{v}_t\cdot x_t(b_t)\right] &\leq   \E_{\vec{\gamma}\sim \vec{\mathcal{P}}}\left[\left(1+\frac{d_{\text{max}}}{v_{\text{min}}}\right)\sum_{t=1}^T {v}_t \cdot x_t(b_t) \right]\\
         &= \E_{\vec{\gamma}\sim \vec{\mathcal{P}}}\left[\left(1+\frac{d_{\text{max}}}{v_{\text{min}}}\right)\text{Reward}(\text{Alg}, \vec{\gamma})\right].
\end{split}
\end{equation}

Combining \eqref{new_36} and \eqref{new_37}, we have that 
\begin{equation*}
\begin{split}
         &\E_{\vec{\gamma}\sim \vec{\mathcal{P}}}\left[\sum_{t=1}^T \hat{v}_t\cdot x_t(b^*_t) -\sum_{t=1}^T \hat{v}_t\cdot x_t(b_t)\right]\\
         \geq&\E_{\vec{\gamma}\sim \vec{\mathcal{P}}}\left[\text{Reward}(\vec{b}^{\text{OPT}}, \vec{\gamma})\right]- \E_{\vec{\gamma}\sim \vec{\mathcal{P}}}\left[\left(1+\frac{d_{\text{max}}}{v_{\text{min}}}\right)\text{Reward}(\text{Alg}, \vec{\gamma})\right].
\end{split}
\end{equation*}

This completes the proof of Theorem \ref{theo_1}.
\end{proof}

\subsection{Proof of Theorem \ref{thoe:3}}\label{A3.1}
\begin{proof}~
\noindent
From the definitions of $d_{\text{max}}$, we obtain the following inequality
\begin{equation}\label{app_24}
\begin{split}
          \frac{ \sum_{t=1}^T ({v}_t-d_{\text{max}}) \cdot x_t(b_t)}{\sum_{t=1}^T {v}_t \cdot x_t(b_t) }&\leq \frac{ \sum_{t=1}^T \hat{v}_t \cdot x_t(b_t)}{\sum_{t=1}^T {v}_t \cdot x_t(b_t) }.  
\end{split}
\end{equation}

From the definitions of $ {v}_{\text{min}} $,, we obtain
\begin{equation}\label{app_25}
    \frac{ \sum_{t=1}^T ({v}_t-d_{\text{max}}) \cdot x_t(b_t)}{\sum_{t=1}^T {v}_t \cdot x_t(b_t) }\geq 1-\frac{ d_{\text{max}} \cdot\sum_{t=1}^T  x_t(b_t)}{ {v}_{\text{min}} \cdot\sum_{t=1}^T x_t(b_t) }=1-\frac{d_{\text{max}}}{v_{\text{min}}}.
\end{equation}

Thus, combining \eqref{app_22}, \eqref{app_24}, and \eqref{app_25}, we obtain the following inequality
\begin{equation}\label{12}
   1-\frac{d_{\text{max}}}{v_{\text{min}}} \leq \frac{ \sum_{t=1}^T \hat{v}_t \cdot x_t(b_t)}{\sum_{t=1}^T {v}_t \cdot x_t(b_t) } \leq1+\frac{d_{\text{max}}}{v_{\text{min}}}.
\end{equation}

Since 
\begin{equation*}
   \widehat{\text{RoS}}(\vec{b}, \vec{\gamma})={\text{RoS}}(\vec{b}, \vec{\gamma}) \cdot \frac{ \sum_{t=1}^T \hat{v}_t \cdot x_t(b_t)}{\sum_{t=1}^T {v}_t \cdot x_t(b_t) },
\end{equation*}
combining \eqref{12}, we get 
\begin{equation*}
\left (1-\frac{d_{\text{max}}}{v_{\text{min}}}\right)\text{RoS}(\vec{b}, \vec{\gamma})\leq \widehat{\text{RoS}}(\vec{b}, \vec{\gamma})\leq \left (1+\frac{d_{\text{max}}}{v_{\text{min}}}\right)\text{RoS}(\vec{b}, \vec{\gamma}).  
\end{equation*}

This completes the proof of Theorem \ref{thoe:3}.
\end{proof}

\subsection{Proof of Proposition \ref{prop2}}\label{Ap.2}
\begin{proof}~
\noindent By the definitions of $d_{\text{max}}$ and $d_t$, we have:
\begin{equation*}
    \begin{split}
        &d_{\text{max}} = \max_{t \in [T]} \left( |\hat{\mu}(z_t) + d_t - v_t| \right) \\
        \leq& \max_{t \in [T]} \left( |\hat{\mu}(z_t)- v_t  | \right)+\sup_{(z,v) }|\hat{\mu}(z)-v|\\
        \leq& 2\cdot\sup_{(z,v) }|\hat{\mu}(z)-v|.\\
    \end{split}
\end{equation*}

From the model \eqref{model}, we have:
\begin{equation*}
\begin{split}
       \sup_{(z,v) }|\hat{\mu}(z)-v|&=\sup_{(z,v)}|\hat{\mu}(z)-\mu(z)-(v-\mu(z))|\\
   &\leq\sup_{z}|\hat{\mu}(z)-\mu(z)|+\sup_{(z,v)}|v-\mu(z)|,
\end{split}
\end{equation*}
Here, $\sup_{z} |\hat{\mu}(z) - \mu(z)|$ denotes the prediction error of the machine learning algorithm, and $\sup_{(z,v)} |v - \mu(z)|$ represents the value approximation error. The supremum is taken over any pair of feature and value $(z, v)$ in the space $\mathcal{Z} \times [0,1]$.

This completes the proof of Proposition \ref{prop2}.
\end{proof}

\subsection{Proof of Theorem \ref{theo_3}}\label{A4.0}
\begin{proof}~
\noindent The proof of this result follows the framework outlined in \citet{feng2023online}, with extensions to the non-i.i.d. setting based on \citet{balseiro2023best} and incorporating the use of adjusted predicted values.
We now proceed with the full proof of the reward analysis in our setting. First, we define the following terms.

   \begin{definition}We need the following technical definitions. 
\begin{itemize}[leftmargin=*]
\item For some $\lambda\geq 0$ and $\mu \geq 0$,  let $f_t(b):= \hat{v}_t \cdot x_t(b)$, $g_t:=\hat{v}_t\cdot x_t(b)-p_t(b)$, and define
\begin{equation}\label{eq:1-combined}
f^*_t(\mu, \lambda):=\max_{b}\left[f_t(b)+\lambda\cdot g_t(b) -\mu\cdot p_t(b)\right].
\end{equation}
\item  The following dual variable parametrized by $\mathcal{P}$; the quantity $f^*$ is defined in the same way as in \eqref{eq:1-combined}:  
\begin{equation*}
\bar{\mathcal{D}}(\mu, \lambda|\mathcal{P}):=\mu \cdot \frac{B}{T} + \E_{(z, v,x,p)\sim\mathcal{P}}\left[f^*(\mu,\lambda)\right].\label{eq:2-combined}
\end{equation*} 
\end{itemize}
\end{definition}

\begin{definition}
    The stopping time $\tau$ of Algorithm \ref{alg:1}, with a total initial budget of $B$ is the first time $\tau$ at which $\sum^{\tau}_{t=1} p_t(b_t) +1 \geq B.$
    Intuitively, this is the first time step at which the total price paid almost exceeds the total budget. 
\end{definition}

The proof proceeds in three main steps.

\textbf{Step 1. Lower Bound of the Obtained Values}

Until the stopping time $t=\tau$, we have that \[ f^*_t(\mu_t, \lambda_t)=\hat{v}_t \cdot x_t(b_t)+\lambda_t\cdot g_t(b_t)-\mu_t\cdot p_t(b_t).\]
Rearranging the terms and taking the expectation conditioned on the randomness up to step $t-1$, we obtain,
\begin{equation}\label{e17}
\E_{\vec{\gamma}\sim \vec{\mathcal{P}}}[\hat{v}_t \cdot x_t(b_t)|\vec{\gamma}_{-t}]=\E_{\vec{\gamma}\sim \vec{\mathcal{P}}}[f^*_t(\mu_t, \lambda_t)+\mu_t\cdot p_t(b_t)-\lambda_t\cdot g_t(b_t)|\vec{\gamma}_{-t}],
\end{equation}
where $\vec{\gamma}_{-t}=(\gamma_1,\dots,\gamma_{t-1})$.

From Algorithm \ref{alg:1}, once the randomness up to step $t-1$ is fixed, the dual variables are also fixed. Therefore, we have
\begin{equation}\label{e18}
\begin{split}
    &\E_{\vec{\gamma}\sim \vec{\mathcal{P}}}[f^*_t(\mu_t, \lambda_t)|\vec{\gamma}_{-t}]=\bar{\mathcal{D}}(\mu_t, \lambda_t|\mathcal{P}_t)-\mu_t \cdot ({B}/{T}).\\
\end{split}
\end{equation}
Combining equations \eqref{e17} and \eqref{e18}, we obtain
\begin{equation*}
\begin{split}
        &\E_{\vec{\gamma}\sim \vec{\mathcal{P}}}[\hat{v}_t \cdot x_t(b_t)|\vec{\gamma}_{-t}]\\
    =&\bar{\mathcal{D}}(\mu_t, \lambda_t|\mathcal{P}_t)- \E_{\vec{\gamma}\sim \vec{\mathcal{P}}}[\mu_t\cdot\left({B}/{T}-p_t(b_t)\right)+\lambda_t\cdot g_t(b_t)|\vec{\gamma}_{-t}].
\end{split}
\end{equation*}

Summing over $t=1,2,\dots, \tau$, and using the Optional Stopping Theorem, we get 
\begin{equation}\label{19}
\begin{split}
         \E_{\vec{\gamma}\sim \vec{\mathcal{P}}}\left[\sum_{t=1}^{\tau}\hat{v}_t \cdot x_t(b_t)\right]=
     \E_{\vec{\gamma}\sim \vec{\mathcal{P}}}\left[\sum_{t=1}^{\tau}(\bar{\mathcal{D}}(\mu_t, \lambda_t|\mathcal{P}_t))\right]-\\\E_{\vec{\gamma}\sim \vec{\mathcal{P}}}\left[\sum_{t=1}^{\tau}\mu_t\cdot\left({B}/{T}-p_t(b_t)\right)+\sum_{t=1}^{\tau}\lambda_t\cdot g_t(b_t)\right].
\end{split}
\end{equation}

Define $\bar{\mathcal{P}}=\frac{1}{T}\sum_{s=1}^T\mathcal{P}_s$. Then, we have that
\begin{equation}\label{e20}
\begin{split}
    &\sum_{t=1}^{\tau}(\bar{\mathcal{D}}(\mu_t, \lambda_t|\mathcal{P}_t))
    =\sum_{t=1}^{\tau}(\E_{\gamma_t \sim\mathcal{P}_t}\left[f_t^*(\mu_t,\lambda_t)\right]+\mu_t \cdot ({B}/{T}))\\
    =&\sum_{t=1}^{\tau}(\E_{\gamma_t \sim\bar{\mathcal{P}}}\left[f_t^*(\mu_t,\lambda_t)\right]+\mu_t \cdot ({B}/{T}))\\
     &+\sum_{t=1}^{\tau}(\E_{\gamma_t \sim\mathcal{P}_t}\left[f_t^*(\mu_t,\lambda_t)\right]-\E_{\gamma_t \sim\bar{\mathcal{P}}}\left[f_t^*(\mu_t,\lambda_t)\right])\\
    =&\sum_{t=1}^{\tau}(\bar{\mathcal{D}}(\mu_t, \lambda_t|\bar{\mathcal{P}}))+\sum_{t=1}^{\tau}(\E_{\gamma_t \sim\mathcal{P}_t}\left[f_t^*(\mu_t,\lambda_t)\right]-\E_{\gamma_t \sim\bar{\mathcal{P}}}\left[f_t^*(\mu_t,\lambda_t)\right])\\
    \geq&\tau \bar{\mathcal{D}}(\bar{\mu}_{\tau}, \bar{\lambda}_{\tau}|\bar{\mathcal{P}})- (\max_{1\leq t\leq \tau}(\lambda_t \cdot g_t(b_t))+1)\text{MD}(\vec{\mathcal{P}}),
\end{split}
\end{equation}
where $\bar{\mu}_{\tau}=\frac{1}{\tau}\sum^{\tau}_{t=1}\mu_t$, $\bar{\lambda}_{\tau}=\frac{1}{\tau}\sum^{\tau}_{t=1}\lambda_t$, and the inequality follows from the convexity of the dual function, as well as the fact that $f^*_t(\mu, \lambda)\in [0,\max_{1\leq t\leq \tau}(\lambda_t \cdot g_t(b_t))+1]$.

Combining \eqref{19} and \eqref{e20}, we obtain
\begin{equation}\label{e23}
\begin{split}
         &\E_{\vec{\gamma}\sim \vec{\mathcal{P}}}\left[\sum_{t=1}^{\tau}\hat{v}_t \cdot x_t(b_t)\right]
         \geq \E_{\vec{\gamma}\sim \vec{\mathcal{P}}} [\tau \bar{\mathcal{D}}(\bar{\mu}_{\tau}, \bar{\lambda}_{\tau}|\bar{\mathcal{P}})] \\
         &-\E_{\vec{\gamma}\sim \vec{\mathcal{P}}}\left[\sum_{t=1}^{\tau}\mu_t\cdot\left({B}/{T}-p_t(b_t)\right)+\sum_{t=1}^{\tau}\lambda_t\cdot g_t(b_t)\right]\\
         & -(\max_{1\leq t\leq \tau}(\lambda_t \cdot g_t(b_t))+1)\text{MD}(\vec{\mathcal{P}}).
\end{split}
\end{equation}

\begin{table*}[ht!]
\centering
\setlength\tabcolsep{4pt} 
\caption{Average computation time (in seconds) of different algorithms across various periods.}
{%
\begin{tabular}{ccccccccc}
\toprule 

\( \textbf{Method} \) & P8 & P9 & P10 & P11 & P12 & P13 & P14 & AVG $\pm$ STD \\  \midrule
UCB  & 2914.8 & 5208.5 & 3212.3 & 1485.4 & 2583.3 & {6935.0} & 5407.9 & 3985.7 $\pm$ 1579.0 \\  
Pred  & 2.1 & 2.3 & 2.2 & 2.1 & 2.2 & {2.5} & 2.4 &  2.3 $\pm$ 0.1 \\ 
Adjust  & {3.6} & {3.9} & {3.4} & {4.6} & {4.6} & {5.2} & {4.5} & 4.3 $\pm$ 0.7\\

\bottomrule 
\end{tabular}%
}
\label{tab:time_ali} 
\end{table*}

\begin{table*}[!ht]
\centering
\setlength\tabcolsep{4pt} 
\caption{Standard deviation of the results in Table \ref{tab:per_ali} across all bidders.}
{%
\begin{tabular}{ccccccccc}
\toprule 

\( \textbf{Method} \) & P8 & P9 & P10 & P11 & P12 & P13 & P14 & AVG $\pm$ STD \\  \midrule
UCB  & 0.158 & 0.146 & 0.158 & 0.260 & 0.230 & {0.121} & 0.142 & 0.174 $\pm$ 0.055  \\  
Pred  & 0.050 & 0.049 & 0.030 & 0.015 & 0.030 & {0.028} & 0.004 & 0.028 $\pm$ 0.018\\ 
Adjust  & {0.049} & {0.049} & {0.033} & {0.026} & {0.036} & {0.032} & {0.017} & 0.032 $\pm$ 0.010\\

\bottomrule 
\end{tabular}%
}
\label{tab:sd_ali} 
\end{table*}

\begin{table*}
\centering
\setlength\tabcolsep{4pt} 
\caption{Standard deviation of the results in Table \ref{tab:per_jd} across all bidders for different numbers of historical data $n$ and new auctions $T$.}
{%
\begin{tabular}{cccccccccc}
\toprule 
& \multicolumn{3}{c}{\( n = 6000 \)} & \multicolumn{3}{c}{\( n = 8000 \)} & \multicolumn{3}{c}{\( n = 10000 \)} \\ 
\cmidrule(l{0.75em}r{0.75em}){2-4}
\cmidrule(l{0.75em}r{0.75em}){5-7}
\cmidrule(l{0.75em}r{0.75em}){8-10}
\( T \) & UCB & Pred & Adjust & UCB & Pred & Adjust & UCB & Pred & Adjust  \\  \midrule
1500  & 1.34 & 0.06 & 0.04 & 1.38 & 0.06 & 0.05 & 1.28 & 0.05 & 0.04 \\  
1900  & 1.19 & 0.06 & 0.05 & 1.28 & 0.06 & 0.05 & 1.29 & 0.06 & 0.05\\ 
2300  & 1.34 & 0.06 & 0.05 & 1.24 & 0.05 & 0.05 & 1.28 & 0.06 & 0.05\\
2700  & 1.31 & 0.06 & 0.05 & 1.35 & 0.06 & 0.05 & 1.27 & 0.06 & 0.05\\
\bottomrule 
\end{tabular}%
}
\label{tab:sd_jd} 
\end{table*}

\textbf{Step 2. Upper Bound on the Offline Optimum without Knowledge of the Bidder's True Values}

By definition, $(\{b_{t}^{*}\}_{t=1}^T, \{\hat{v}_t\}_{t=1}^T)$ represents the optimal solution to Problem \eqref{prob_2}, and therefore,
\begin{equation*}
\begin{split}
    &\sum_{t=1}^T \hat{v}_t\cdot x_t(b_t^*)=\max_{\{b_t\}:\sum_{t=1}^T p_t(b_t)\leq B, \sum_{t=1}^T g_t(b_t)\geq 0 }\left[ \sum_{t=1}^T \hat{v}_t\cdot x_t(b_t) \right]\\
    =&\max_{\{b_t\}}\min_{\lambda\geq 0, \mu \geq 0}\left[\sum_{t=1}^T   \hat{v}_t\cdot x_t(b_t)+\lambda \cdot \sum_{t=1}^T g_t(b_t)+\mu\cdot (B-\sum_{t=1}^T p_t(b_t))\right].
\end{split}
\end{equation*}

By applying Sion's minimax theorem and the definition of $f^*_t$, we obtain
\begin{equation*}
    \begin{split}
         &\sum_{t=1}^T \hat{v}_t\cdot x_t(b_t^*)\leq  \min_{\lambda\geq 0, \mu \geq 0} \sum_{t=1}^T \mu\cdot \frac{B}{T} 
         \\
         +& \max_{b_t}[\hat{v}_t\cdot x_t(b_t)+\lambda\cdot g_t(b_t)-\mu\cdot p_t(b_t)]\\
         =&\min_{\lambda\geq 0, \mu \geq 0} \sum_{t=1}^T[ \mu\cdot \frac{B}{T} +f^*_t(\mu,\lambda)]\leq \sum_{t=1}^T\left[ \mu\cdot \frac{B}{T}+f^*_t(\bar{\mu}_{\tau},\bar{\lambda}_{\tau})\right].
    \end{split}
\end{equation*}
Thus, we have
\begin{equation*}
    \begin{split}
         \E_{\vec{\gamma}\sim \vec{\mathcal{P}}}\left[ \sum_{t=1}^T \hat{v}_t\cdot x_t(b_t^*)\right]&\leq \sum_{t=1}^T \left(\mu\cdot \frac{B}{T}+\E_{\gamma_t \sim\mathcal{P}_t}\left[ f^*_t(\bar{\mu}_{\tau},\bar{\lambda}_{\tau})\right]\right)\\
         &=T\cdot\bar{\mathcal{D}}(\bar{\mu}_{\tau},\bar{\lambda}_{\tau}|\bar{\mathcal{P}}),
    \end{split}
\end{equation*}
where the second equation follows because $\bar{\mathcal{P}}$ is the time-averaged distribution of requests.

We can then derive the following
\begin{equation}\label{e28}
     \begin{split}
        &\E_{\vec{\gamma}\sim \vec{\mathcal{P}}}\left[ \sum_{t=1}^T \hat{v}_t\cdot x_t(b_t^*)\right] \\
        =& \frac{\tau}{T} \cdot \E_{\vec{\gamma}\sim \vec{\mathcal{P}}}\left[ \sum_{t=1}^T \hat{v}_t\cdot x_t(b_t^*)\right]+\frac{T-\tau}{T}\cdot\E_{\vec{\gamma}\sim \vec{\mathcal{P}}}\left[ \sum_{t=1}^T \hat{v}_t\cdot x_t(b_t^*)\right]\\
        \leq& \tau \bar{\mathcal{D}}(\bar{\mu}_{\tau},\bar{\lambda}_{\tau}|\bar{\mathcal{P}})+(T-\tau),
    \end{split}
\end{equation}
where the inequality follows from the fact $\E_{\vec{\gamma}\sim \vec{\mathcal{P}}}\left[ \sum_{t=1}^T \hat{v}_t\cdot x_t(b_t^*)\right]\leq T$.

\textbf{Step 3. Putting It All Together.}

From \eqref{e23} and \eqref{e28}, we have that 
\begin{equation}\label{e29}
\begin{split}
        &\E_{\vec{\gamma}\sim \vec{\mathcal{P}}}\left[\sum_{t=1}^T \hat{v}_t\cdot x_t(b^*_t)\right]-\E_{\vec{\gamma}\sim \vec{\mathcal{P}}}\left[\sum_{t=1}^T \hat{v}_t\cdot x_t(b_t)\right]\\
        \leq & (T-\tau)+ \E_{\vec{\gamma}\sim \vec{\mathcal{P}}}\left[\sum_{t=1}^{\tau}\mu_t\cdot\left({B}/{T}-p_t(b_t)\right)+\sum_{t=1}^{\tau}\lambda_t\cdot g_t(b_t)\right]\\
         &\quad  +(\max_{1\leq t\leq \tau}(\lambda_t \cdot g_t(b_t))+1)\text{MD}(\vec{\mathcal{P}}).
        \end{split}
\end{equation}

By \citet{feng2023online}, we have
\begin{equation}\label{e30}
\sum_{t=1}^{\tau}\lambda_t\cdot g_t(b_t)\leq O(\sqrt{T})
\end{equation}
and 
\begin{equation}\label{app_36}
\sum_{t=1}^{\tau}\mu_t\cdot\left({B}/{T}-p_t(b_t)\right)\leq (\tau -T)+\frac{T}{B} +O(\sqrt{T}).
\end{equation}

From  \eqref{e30}, it holds that 
\begin{equation}\label{app_37}
  \max_{1\leq t\leq \tau}(\lambda_t\cdot g_t(b_t))+1    \leq O(\sqrt{T}).  
\end{equation}

Substituting all the three inequalities \eqref{e30}, \eqref{app_36}, and \eqref{app_37} into equation \eqref{e29}, we obtain
\begin{equation}\label{eq39}
    \E_{\vec{\gamma}\sim \vec{\mathcal{P}}}\left[\sum_{t=1}^T \hat{v}_t\cdot x_t(b^*_t) -\sum_{t=1}^T \hat{v}_t\cdot x_t(b_t)\right]\leq \frac{T}{B}+O(\sqrt{T})(1+\text{MD}(\vec{\mathcal{P}})).
\end{equation}

Combining Theorem \ref{theo_1} and \eqref{eq39}, and under the conditions $\frac{T}{B}=O(1)$ and $\text{MD}(\vec{\mathcal{P}})=O(1)$, we obtain 
\begin{equation*}
    \E_{\vec{\gamma}\sim \vec{\mathcal{P}}}\left[ \frac{1}{1+\xi}\text{Reward}(\vec{b}^{\text{OPT}}, \vec{\gamma})-\text{Reward}(\text{Alg}, \vec{\gamma})\right]\leq  O(\sqrt{T}),
\end{equation*}
where $\xi = {d_{\text{max}}}/v_{\text{min}}$.

The budget constraint is never violated, as the agent will only place a bid if the remaining budget exceeds the current payment. For the RoS constraint,  from \eqref{12}, we have that
\begin{equation*}
\begin{split}
      \sum_{t=1}^T \hat{v}_t \cdot x_t(b_t)\leq  \sum_{t=1}^T v_t \cdot x_t(b_t)\cdot\left (1+\frac{d_{\text{max}}}{v_{\text{min}}}\right).
\end{split}
\end{equation*}

As established in \citet{feng2023online}, we have the following bound $     \sum_{t=1}^T p_t(b_t)-\sum_{t=1}^T \hat{v}_t \cdot x_t(b_t)\leq O(\sqrt{T}logT).$
Thus, we obtain that
\begin{equation*}
        \sum_{t=1}^T p_t(b_t)-\sum_{t=1}^T v_t \cdot x_t(b_t)\cdot\left (1+\frac{d_{\text{max}}}{v_{\text{min}}}\right)\leq O(\sqrt{T}logT).
\end{equation*}

This completes the proof of Theorem \ref{theo_3}.
\end{proof}

\section{Additional Numerical Results}
\subsection{Additional Numerical Results on Simulation Study}\label{add_simu}

Table \ref{tab:time_ali} reports the average computation time (in seconds) of different algorithms across various periods, similar to Table \ref{tab:per_ali}. The time cost is measured as wall-clock time on an Intel processor with 32 cores activated for parallel computing, 180 GB of memory, and no GPU acceleration. The results show that the computational cost of UCB is significantly higher, as it requires solving a linear program for each auction. In contrast, our method leverages machine learning predictions to provide uncertainty quantification of the values, offering high performance with minimal additional computational resources, making it more cost-effective for online advertising.

Table \ref{tab:sd_ali} presents the standard deviation of the results in Table \ref{tab:per_ali} across all bidders. The variance of the adjusted value is comparable to that of the predicted results, while UCB exhibits a significantly higher variance, highlighting the stability of our method.

\subsection{Additional Numerical Results on Real World Application}

Table \ref{tab:sd_jd} presents the standard deviation of the results in Table \ref{tab:per_jd} across all bidders for different numbers of historical data $n$ and new auctions $T$. The variance of the adjusted value is comparable to that of the predicted results, while UCB exhibits a significantly higher variance, highlighting the stability of our method.
\end{document}